\documentclass[10pt,twocolumn,letterpaper]{article}

\usepackage[pagenumbers]{cvpr} 

\usepackage{graphicx}
\usepackage{amsmath}
\usepackage{amssymb}
\usepackage{booktabs}

\usepackage{multirow}
\usepackage{pifont}
\usepackage{mathrsfs}
\usepackage{multirow,overpic}

\usepackage{verbatim}
\usepackage{color}
\usepackage{float}
\usepackage{enumitem}
\usepackage{booktabs}
\usepackage{tabulary,multirow,overpic,xcolor}

\usepackage[pagebackref,breaklinks,colorlinks]{hyperref}

\usepackage[capitalize]{cleveref}
\crefname{section}{Sec.}{Secs.}
\Crefname{section}{Section}{Sections}
\Crefname{table}{Table}{Tables}
\crefname{table}{Tab.}{Tabs.}

\usepackage{xcolor}
\usepackage[font=small,labelfont=bf]{caption}  
\usepackage{makecell}
\usepackage{tabulary}
\definecolor{demphcolor}{RGB}{144,144,144}
\newcommand{\demph}[1]{\textcolor{demphcolor}{#1}}
\definecolor{mygray}{gray}{0.4}
\usepackage{pifont}
\newcommand{\cmark}{\color{mygray}\ding{51}}%
\newcommand{\xmark}{\color{mygray}\ding{55}}%
\newlength\savewidth\newcommand\shline{\noalign{\global\savewidth\arrayrulewidth
  \global\arrayrulewidth 1pt}\hline\noalign{\global\arrayrulewidth\savewidth}}

\newcommand{\tablestyle}[2]{\setlength{\tabcolsep}{#1}\renewcommand{\arraystretch}{#2}\centering\footnotesize}
\makeatletter\renewcommand\paragraph{\@startsection{paragraph}{4}{\z@}
  {.5em \@plus1ex \@minus.2ex}{-.5em}{\normalfont\normalsize\bfseries}}\makeatother
\newcolumntype{C}[1]{>{\centering\arraybackslash}p{#1}}
\newcolumntype{R}[1]{>{\raggedleft\arraybackslash}p{#1}}
\newcolumntype{L}[1]{>{\raggedright\arraybackslash}p{#1}}

\newcommand{\specialcelll}[2][l]{%
  \begin{tabular}[#1]{@{}l@{}}#2\end{tabular}}

\newcommand{\Paragraph}[1]{\vspace{1mm} \noindent \textbf{#1} \hspace{0mm}}

\definecolor{asparagus}{rgb}{0.53, 0.66, 0.42}
\definecolor{blue(munsell)}{rgb}{0.0, 0.5, 0.69}

\newcommand{\kevinarxiv}[1]{\textcolor{black}{#1}}
\newcommand*\samethanks[1][\value{footnote}]{\footnotemark[#1]}
\def\cvprPaperID{3770} 
\def\confName{CVPR}
\def\confYear{2022}

\begin{document}


\title{\textsc{SwinBERT:} End-to-End Transformers with Sparse Attention  \\for Video Captioning}


\author{Kevin Lin\thanks{\ Equal contribution.}, Linjie Li\samethanks, Chung-Ching Lin\samethanks, Faisal Ahmed, Zhe Gan, Zicheng Liu, Yumao Lu, Lijuan Wang\\
Microsoft\\
{\tt\small \{keli, lindsey.li, chungching.lin, fiahmed, zhe.gan, zliu, yumaolu, lijuanw\}@microsoft.com}
}

\maketitle

\begin{abstract}
   The canonical approach to video captioning dictates a caption generation model to learn from offline-extracted dense video features. These feature extractors usually operate on video frames sampled at a fixed frame rate and are often trained on image/video understanding tasks, without adaption to video captioning data. In this work, we present \textsc{SwinBERT}, an end-to-end transformer-based model for video captioning, which takes video frame patches directly as inputs, and outputs a natural language description. Instead of leveraging multiple 2D/3D feature extractors,  our method adopts a video transformer to encode spatial-temporal representations that can adapt to variable lengths of video input without dedicated design for different frame rates. Based on this model architecture, we show that video captioning can benefit  significantly from more densely sampled video frames as opposed to previous successes with sparsely sampled video frames for video-and-language understanding tasks (\textit{e.g.}, video question answering). Moreover, to avoid the inherent redundancy in consecutive video frames, we propose adaptively learning a sparse attention mask and optimizing it for task-specific performance improvement through better long-range video sequence modeling. Through extensive experiments on 5 video captioning datasets, we show that \textsc{SwinBERT} achieves across-the-board performance improvements over previous methods, often by a large margin. The learned sparse attention masks in addition push the limit to new state of the arts, and can be transferred between different video lengths and between different datasets. Code is available at \url{https://github.com/microsoft/SwinBERT}.
\end{abstract}


\section{Introduction}\label{sec:intro}
\begin{figure}[t!]
\begin{center}
\includegraphics[trim=0 0 0 0, clip,width=1\linewidth]{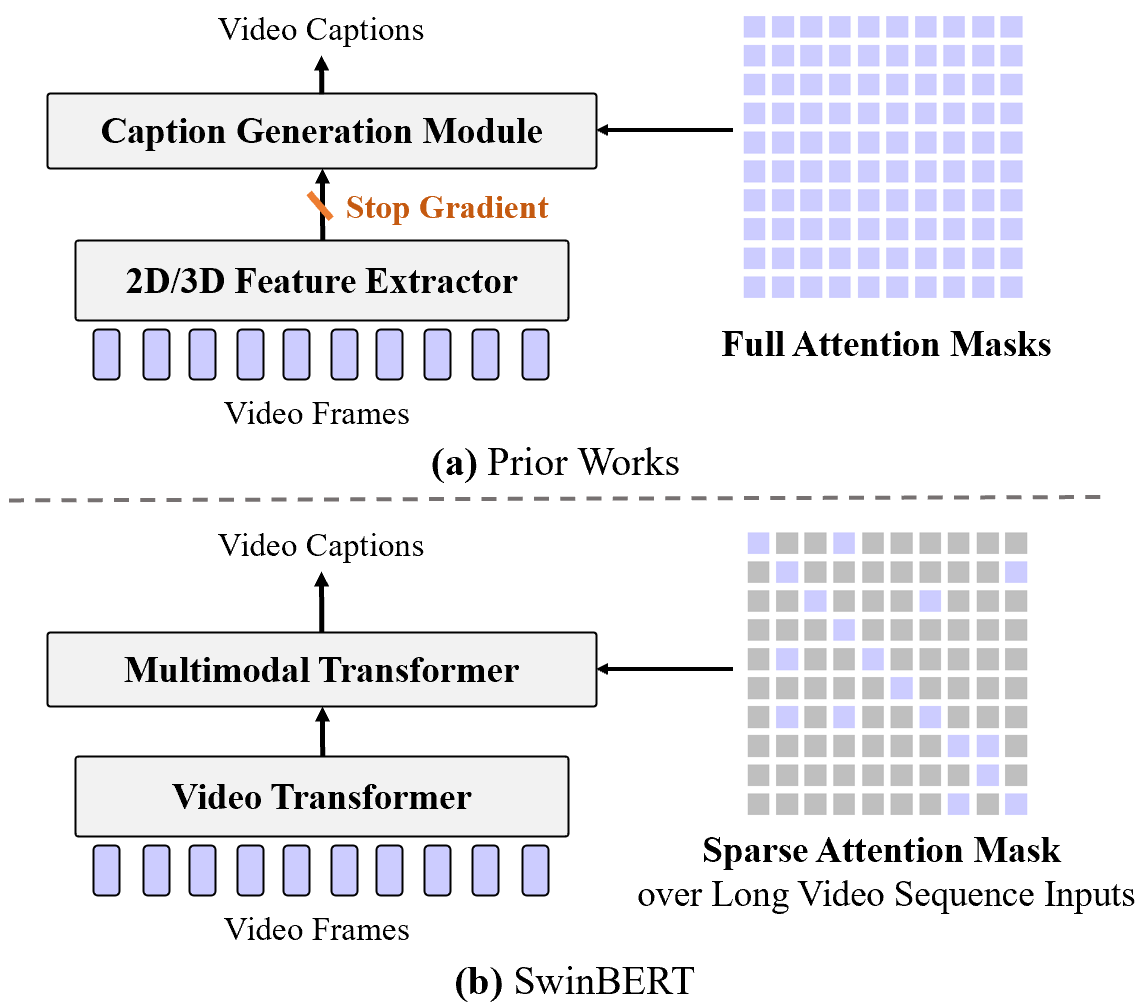}
\vspace{-7mm}
\caption{\small
Comparison between previous works and \textsc{SwinBERT}. 
Different from prior works that use offline-extracted 2D/3D features, we propose to adopt the video transformer as our video encoder, and present an end-to-end fully Transformer-based model for video captioning.
We further propose to adaptively learn a sparse attention mask to improve long-range video sequence modeling. 
}
\vspace{-10mm}
\label{fig:intro}
\end{center}
\end{figure}

Video captioning~\cite{ijcai2019-877,liu2020sibnet, zhang2021open, VALUE, shi2020learning,pan2020spatio,aafaq2019spatio, pei2019memory, sun2019videobert} is the task of describing the visual content of a given video in natural language.
As such, it requires
an algorithm to understand and model the spatial-temporal dynamics in video, as well as the relationships between visual and textual elements, and to generate a sequence of output words. This has usually been tackled with transformer-based models that learn from offline extracted video representations~\cite{VALUE,lei2021less,luo2020univl,sun2019videobert} (Figure~\ref{fig:intro} (a)). Specifically, multiple feature extractors, usually trained on image/video understanding tasks (\textit{e.g.}, image classification or action recognition), are employed to extract 2D appearance features and 3D motion features from densely sampled video frames. Although achieving promising results, there exists a discrepancy in both data domain and task formulation between these off-the-shelf feature extractors and downstream video captioning. However, end-to-end training with multiple feature extractors on such dense video frames is computationally intensive, or even infeasible. 

More recently, \textsc{ClipBERT}~\cite{lei2021less} points out the repetitive information presented in consecutive video frames is not necessary for downstream video-and-language tasks, and proposes a sparse sampling strategy that enables affordable 
end-to-end training to the raw pixel inputs. Although it has shown great success in video-and-language understanding tasks, such as video question answering~\cite{lei2018tvqa} and text-to-video retrieval~\cite{lei2020tvr,xu2016msr}, it remains unclear whether these sparsely sampled video frames are sufficient to generate rich and descriptive captions. Moreover, \textsc{ClipBERT} leverages a 2D Convolutional Neural Network together with mean pooling that operates directly on the raw video frames to learn video representations, which may lose temporal information that is essential to describe visual events in chronological order. 

In this work, we aim to find an end-to-end solution to the video captioning task.
Inspired by the recent successes of Transformer-based models in computer vision~\cite{dosovitskiy2020image,arnab2021vivit,liu2021video,bertasius2021space}, especially for video understanding tasks~\cite{carreira2017quo}, 
we propose \textsc{SwinBERT} (Figure~\ref{fig:intro} (b)), a pure Transformer-based model that directly takes raw video frames as inputs for end-to-end caption generation. Unlike previous methods leveraging off-the-shelf 2D/3D feature extractors at a fixed frame rate, we employ a video Transformer capable of learning from variable lengths of video frame sequence  without dedicated design for different frame rates.  
Based on this specific model design,  we investigate \emph{how many video frames are sufficient for the video captioning task?}. Our experiments show that the captioning performance (\textit{i.e.}, CIDEr score) can be greatly lifted by more densely sampled frames (\textit{e.g.}, Ours: 64 frames, vs. \textsc{ClipBERT}: 16 frames), in contrast to previous success with sparsely sampled frames for video-and-language understanding tasks. 
Lastly, to avoid the redundancy that comes naturally in consecutive video frames, we further introduce a learnable Sparse Attention Mask as a regularizer that allows the model to focus more on video frame patches that contain more spatial-temporal movements (\textit{e.g.}, the main moving objects) than those staying unchanged for the entire video duration (\textit{e.g.}, the background). 
In contrast to prior models~\cite{VALUE,lei2021less,luo2020univl} with predefined attention structures, our model can learn adaptive  attention maps to optimize for task-specific performance improvements through better video sequence modeling.

\begin{table}[t]
\tablestyle{5pt}{1.1} 
\def\w{20pt} 
\centering
\resizebox{\linewidth}{!}{
\begin{tabular}{lcccccc}
    \toprule
	Method  & MSVD $\uparrow$ & YouCook2 $\uparrow$ & MSRVTT $\uparrow$ & TVC $\uparrow$ & VATEX $\uparrow$ \\
	\midrule
    SOTA  & $95.2$~\cite{Zhang_2020_CVPR} & $53.6$~\cite{VALUE} & $52.9$~\cite{zhang2021open} & $51.0$~\cite{VALUE} & $58.1$~\cite{VALUE}\\
    \textsc{SwinBERT}  & $\kevinarxiv{\textbf{120.6}}$ & $\textbf{109.0}$ & $\kevinarxiv{\textbf{53.8}}$ & $\textbf{56.9}$ & $\textbf{73.0}$\\
	\bottomrule
\end{tabular}
}
\vspace{-2mm}
\caption{Comparison with state-of-the-art methods across all video captioning datasets considered on  CIDEr~\cite{vedantam2015cider} metric.}
\vspace{-2mm}
\label{tbl:compare-to-sota}
\end{table}

Our extensive experimental results on 5 video captioning datasets demonstrate that our proposed model is effective in learning sparse attention patterns to improve long-range video sequence modeling, and consequently outperforms previous state-of-the-art approaches by a large margin. To the best of our knowledge, \textsc{SwinBERT} is the first end-to-end pure Transformer-based architecture for video captioning. Additionally, the proposed Sparse Attention Mask effectively regularizes model training and brings further performance improvements across all 5 datasets, which opens a new direction in removing redundancy in video inputs for video-and-language modeling. 


In summary, our contributions are three-fold.
\begin{itemize}
\item{We present \textsc{SwinBERT}, the first end-to-end fully Transformer-based model for video captioning.}
\vspace{-1.5mm}
\item{We introduce the Sparse Attention Mask as a regularizer for improving long-range video sequence modeling, and quantitatively validate the effectiveness of the learnable sparse attention mask in caption generation.}
\vspace{-1.5mm}
\item{Our method outperforms previous state-of-the-art methods by a large margin on 5 popular video captioning benchmarks. As shown in Table~\ref{tbl:compare-to-sota}, \textsc{SwinBERT} achieves an absolute CIDEr improvement of \kevinarxiv{+25.4} on MSVD, +55.4 on YouCook2,  \kevinarxiv{+0.9} on MSRVTT,  +5.9 on TVC and +14.9 on VATEX.}
\end{itemize}
\section{Related Work}\label{sec:related}

\begin{figure*}[t!]
\begin{center}
\includegraphics[trim=0 0 0 0, clip,width=1\textwidth]{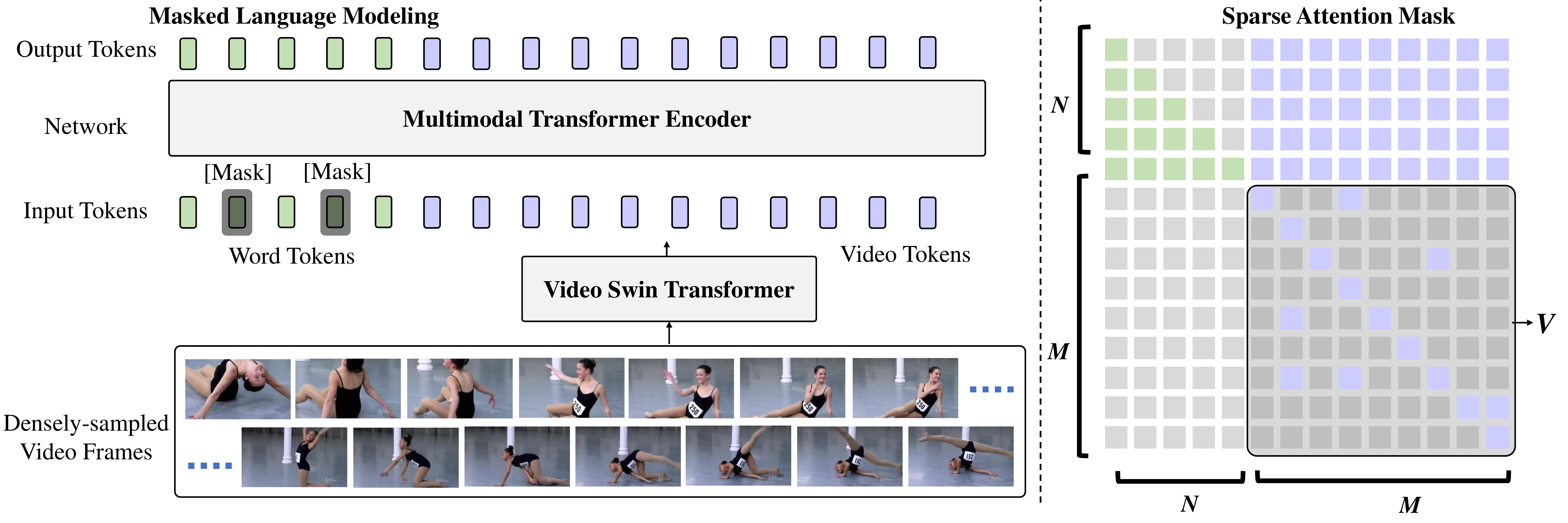}
\vspace{-7mm}
\caption{\small
\textbf{Overview of the proposed framework.} Our model takes a sequence of video frames as inputs, and extracts a set of video tokens using a Video Swin Transformer (VidSwin). Given the word tokens and video tokens, we perform self-attention through multiple layers of a multimodal transformer encoder, and predict the word tokens via masked language modeling. As shown on the right, we propose a learnable Sparse Attention Mask as a regularizer for multimodal transformer encoder to reduce redundancy among the video tokens. During inference, the model takes a testing video sample (single-modality) to generate natural language descriptions. 
}
\vspace{-7mm}
\label{fig:overview}
\end{center}
\end{figure*}

\Paragraph{Video Captioning.} Recent researches~\cite{luo2020univl, shi2020learning,pan2020spatio,aafaq2019spatio,pei2019memory} mainly focus on modeling the relationship between fixed video representations and the output textual descriptions via an encoder-decoder framework for video captioning. Specifically, these methods~\cite{liu2020sibnet, VALUE, luo2020univl, zhang2021open,ijcai2019-877} employ an encoder to refine video representations from a set of fixed video frame features,
and a language decoder operates on top of these refined video representations to learn visual-textual alignment for caption generation. 
Researchers~\cite{VALUE,pan2020spatio,aafaq2019spatio} have focused on exploring different 2D/3D video representations, including IncepResNetV2~\cite{szegedy2017inception}, ResNet~\cite{he2016deep}, CLIP-ViT~\cite{pmlr-v139-radford21a,dosovitskiy2020image}, SlowFast~\cite{feichtenhofer2019slowfast}, C3D~\cite{hara2018can} and S3D~\cite{miech2020end,xie2018rethinking}, for improving video captioning. In addition, object-level representations~\cite{Zhang_2020_CVPR,zhang2019object,hu2019hierarchical} have been explored to enrich captions with fine-grained objects and actions. 
Prior works~\cite{chen2018less} also studied frame selection schemes to capture informative visual inputs.
Unlike previous studies that learn from multiple offline-extracted 2D/3D features with a fixed sampling rate, we introduce Video Swin Transformer~\cite{liu2021video} as the video encoder in our framework to encode spatial-temporal representations from raw video frames. Benefiting from the flexibility of the transformer architecture, our model can learn with variable number of video tokens and can be trained end-to-end.


\Paragraph{Video transformers.} Dosovitskiy~\etal~\cite{dosovitskiy2020image} demonstrate that a pure-transformer based architecture can outperform its convolutional counterparts in ImageNet classification task~\cite{ILSVRC15}. Since then, there has been a growing interest in applying vision transformer (ViT) to the video domain. For example, ViViT~\cite{arnab2021vivit} and TimeSformer~\cite{bertasius2021space} propose a new transformer architecture that can leverage spatial-temporal attention for improving representation learning. Video Swin Transformer (VidSwin)~\cite{liu2021video} further introduces locality inductive bias into the transformer self-attention, and achieves state-of-the-art performance on action recognition benchmark~\cite{carreira2017quo}. While recent studies~\cite{arnab2021vivit,bertasius2021space,liu2021video} mainly focus on developing video transformer architecture for action recognition, video captioning has not been explored along this research direction, which is the focus of this work.


\Paragraph{Video and language.} Recent studies~\cite{VALUE,lei2021less,luo2021clip4clip, zellers2021merlot,miech2019howto100m,miech2020end} have shown great success on multimodal representation learning for video-and-language understanding. Popular downstream tasks include video question answering~\cite{lei2018tvqa}, text-video retrieval~\cite{lei2020tvr,xu2016msr} and video captioning~\cite{wang2019vatex}. Among the literature, Frozen-in-time~\cite{Bain21} is a relevant study that explores pure transformer-based model design, but they focus on text-video retrieval. Specifically, they employ two independent transformer encoders for visual and textual inputs, respectively. Retrieval is conducted by estimating the similarity between the outputs of their visual and textual encoders. With a similar spirit, CLIP4Clip~\cite{luo2021clip4clip} studied using the pre-trained CLIP~\cite{pmlr-v139-radford21a} as a feature extractor for video retrieval. While existing architectures~\cite{Bain21,luo2021clip4clip} are effective for video retrieval, it cannot be directly applied to video captioning, which is the focus of this work. 

\section{Method}\label{sec:method}
In this section, we present \textsc{SwinBERT}, a new video-based pure-Transformer architecture for caption generation. We first detail the model architecture in Section~\ref{sec:model_arch}, then introduce Sparse Attention Mask in Section~\ref{sec:sparse_attn}.

\subsection{Model Architecture}\label{sec:model_arch}

Figure~\ref{fig:overview} shows the overview of the proposed model. \textsc{SwinBERT} takes a sequence of raw video frames as inputs, and then outputs a natural language description describing the input video. \textsc{SwinBERT} consists of two modules: \textit{Video Swin Transformer} (VidSwin), and \textit{Multimodal Transformer Encoder}. First, we leverage VidSwin to extract spatial-temporal video representations from the raw video frames. Then, our Multimodal Transformer Encoder takes as inputs the video representations and outputs a natural language sentence via sequence-to-sequence (seq2seq) generation. We describe each module in detail as below.

\paragraph{Video Swin Transformer.}
As discussed in \cite{donahue2015long, wang2018temporal}, video understanding benefits from long-range temporal modeling. A simple way is to stack a large number of frames to capture long-range structures. However, it would greatly increase the computational cost. Recently, VidSwin~\cite{liu2021video} is designed to leverage the spatial-temporal locality inherent in videos, and achieves a favorable speed-accuracy trade-off. In the first module of our framework, we propose to use VidSwin as our visual encoder to encode the raw video frames as video feature tokens. VidSwin is pre-trained on the Kinetics action recognition task~\cite{carreira2017quo}. 

Given the raw video frames which are of size $T\times H \times W \times 3$, consisting of $T$ frames and each has $H \times W \times 3$ pixels. We feed them to VidSwin, and extract grid features from the last encoder block of VidSwin. The grid features of VidSwin is defined to be of size $\frac{T}{2} \times \frac{H}{32} \times \frac{W}{32} \times 8C$, where $C$ is the channel dimension. We then tokenize the grid features along the channel dimension, resulting in a total of $\frac{T}{2} \times \frac{H}{32} \times \frac{W}{32}$ video tokens. Each token is a $8C-$dim feature vector. After that, we input the video tokens to the multimodal transformer encoder for caption generation.

With our generic design, it enables end-to-end training for video captioning from the raw video frames. Moreover, benefiting from the flexibility of the transformer architecture, our model is able to process variable lengths of video sequences. As we will show in experiments, the caption performance (\textit{i.e.}, CIDEr scores) can be improved with longer video sequence inputs (\textit{i.e.}, densely-sampled video frames).

\paragraph{Multimodal Transformer Encoder.}
In our second module, we use a transformer encoder to generate natural language description. To be specific, it has textual and visual modality inputs, including the tokenized caption description and the video tokens computed from VidSwin. We then perform seq2seq generation to form a natural language sentence. In the same spirit as in image captioning literature~\cite{li2020oscar,hu2021vivo}, we use a causal self-attention mask where a caption token can only attend to the existing output tokens. This effectively simulates a uni-directional seq2seq generation process. In addition, all the textual tokens have full attentions to the video tokens.

\subsection{Learning with Sparse Attention Mask} \label{sec:sparse_attn}
In general, longer inputs across multiple video segments contain more information. 
However, the computational demand of attention are proportional to input length, which limits the number of input frames.
On the other hand, considering the essence of the video properties, the dense-sampling scheme with consecutive video frames contains redundant and perhaps irrelevant information, which may compromise performance. Hence, how to effectively model a long sequence of video tokens is a unique challenge in our proposed framework. We address it by introducing a learnable Sparse Attention Mask as a regularizer to our multimodal transformer encoder.



As shown to the right of Figure~\ref{fig:overview}, the input to the Transformer is split into two parts: $N$ word tokens and $M$ video tokens. The entire attention mask can be defined of size $(N+M) \times (N+M)$, where $N$ is $50$ and $M=\frac{T}{2} \times \frac{H}{32} \times \frac{W}{32}$ in our experiments. We denote $V$ as the learnable attention mask of size $M\times M$ governing the attentions among the video tokens. For more accurate video captioning, we allow the text tokens with unrestricted attention so they can take advantage of visual details. To address the redundancy among the video tokens, we impose the sparsity constraint overlay on top of $V$ by: 
\begin{equation}\begin{aligned}%
\label{eqn:sparse-loss}%
\mathcal{L}_{\textsc{sparse}} = \lambda \times \sum_{i=1}^{M} \sum_{j=1}^{M} \left| V_{i,j}  \right|, 
\end{aligned} 
\end{equation} 
where $\lambda$ is the regularization hyperparameter, and $V_{i,j}$ are the activation values of the learnable attention mask $V$.

During learning, the sparsity constraint will regularize model training to discover the underlying structure of the video sequences.  Through sparse attention, the model learns to strengthen the most important relationships among different tokens by reducing the likelihood of meaningless connections, while focusing more on the active video tokens that contain rich spatial-temporal information. In this way, the model can produce more expressive and descriptive natural language sentences. 

In our implementation, we apply the sigmoid activation function on the sparse attention mask. Therefore, the sparse attention mask consists of continuous activation between $0$ and $1$. As we will show in our experiments, we can realize a binary mask by simply using a threshold of $0.5$.

\paragraph{Training.}

We train \textsc{SwinBERT} in an end-to-end manner by applying Masked Language Modeling ($\mathcal{L}_{\textsc{MLM}}$)~\cite{devlin2019bert} on top of our multimodal transformer encoder. We mask a percentage of word tokens by replacing them with a pre-defined special token \texttt{[MASK]}. We then ask the multimodal transformer to predict the masked ones. In order to predict a masked word token, the model will have to resort to the video tokens and other word tokens. This facilitates cross-modality representation learning to help ground the caption descriptions in the video context. Moreover, we apply the proposed sparsity constraint on the learnable attention mask to enhance the modeling of the video token sequence.

In summary, our loss function includes $\mathcal{L}_{\textsc{MLM}}$~\cite{devlin2019bert} and $\mathcal{L}_{\textsc{sparse}}$, and we train \textsc{SwinBERT} by simply minimizing the sum of them.

\paragraph{Inference.} During inference, our model takes a video sequence as input (single visual modality), and outputs a natural language sentence. We generate the output sentence in an auto-regressive manner. In other words, our model generates one word token at a time, consuming the previously generated tokens as the inputs of the multimodal transformer encoder. We perform generation until our model outputs a pre-defined ending token \texttt{[EOS]} or reaches the maximum output length.
\section{Experiments}\label{sec:exp}

\begin{table*}[t]
\tablestyle{4pt}{1.1} 
\def\w{20pt} 
\centering
    \begin{tabular}{cccccccccccccc}
        \toprule
        ~ & \multicolumn{3}{c}{Features} & ~ & \multicolumn{4}{c}{MSVD} & ~ & \multicolumn{4}{c}{MSRVTT}\\
        \cmidrule{2-4} \cmidrule{6-9} \cmidrule{11-14} Method & 2D Appearance & 3D Motion & Object Detection & ~ &B4 & M & R & C & ~ & B4 & M & R & C \\
        \midrule
        PickNet~\cite{chen2018less} & ResNet152 & - & - & ~ & 52.3 & 33.3 & 69.6 & 76.5 & ~ & 41.3 & 27.7 & 59.8 & 44.1 \\
        SibNet~\cite{liu2020sibnet} & GoogleNet & - & - & ~ & $54.2$ & $34.8$ & $71.7$ & $88.2$ & ~ & $40.9$ & $27.5$ & $60.2$ & $47.5$ \\
        OA-BTG~\cite{zhang2019object} & ResNet200 & - & MaskRCNN & ~ & $56.9$ & $36.2$ & - & $90.6$ & ~ & $41.4$ & $28.2$ & - & $46.9$ \\
        GRU-EVE~\cite{aafaq2019spatio} & IncepResnetV2 & C3D & YOLO & ~ & $47.9$ & $35.0$ & $71.5$ & $78.1$ & ~ & $38.3$ & $28.4$ & $60.7$ & $48.1$ \\
        MGSA~\cite{chen2019motion} & IncepResnetV2 & C3D & - & ~ & 53.4 & 35.0 & - & 86.7 & ~ & 42.4 & 27.6 & - & 47.5 \\
        POS+CG~\cite{wang2019controllable} & IncepResnetV2 & OpticalFlow & - & ~ & 52.5 & 34.1 & 71.3 & 88.7 & ~ & 42.0 & 28.2 & 61.6 & 48.7 \\
        POS+VCT~\cite{hou2019joint} & IncepResnetV2 & C3D & - & ~ & 52.8 & 36.1 & 71.8 & 87.8  & ~ & 42.3 & 29.7 & \textbf{62.8} & 49.1 \\
        SAAT~\cite{zheng2020syntax}  & IncepResnetV2 & C3D & - & ~ & 46.5 & 33.5 & 69.4 & 81.0 & ~ & 39.9 & 27.7 & 61.2 & 51.0 \\
        STG-KD~\cite{pan2020spatio}  & ResNet101 & I3D & FasterRCNN & ~ & 52.2 & 36.9 & 73.9 & 93.0 & ~ & 40.5 & 28.3 & 60.9 & 47.1 \\
        PMI-CAP~\cite{chen2020learning}  & IncepResnetV2 & C3D & - & ~ & 54.6 & 36.4 & - & 95.1 & ~ & 42.1 & 28.7 & - & 49.4 \\
        ORG-TRL~\cite{Zhang_2020_CVPR} & IncepResnetV2 & C3D & FasterRCNN & ~ & 54.3 & 36.4 & 73.9 & 95.2 & ~ & \textbf{43.6} & 28.8 & 62.1 & 50.9 \\
        OpenBook~\cite{zhang2021open} & IncepResnetV2 & C3D & - & ~ & - & - & - & - & ~ & 42.8 & 29.3 & 61.7 & 52.9 \\
        \midrule
        \textsc{SwinBERT}  & \multicolumn{2}{c}{VidSwin} & - & ~ & \textbf{58.2} & \textbf{41.3} & \textbf{77.5} & \textbf{120.6} & ~ & 41.9 & \textbf{29.9} & 62.1 & \textbf{53.8} \\
        \bottomrule
    \end{tabular}
    \vspace{-2mm}
    \caption{Comparison with state-of-the-art methods on \kevinarxiv{the test split of MSVD and MSRVTT.}}
    \label{table:msvd-msrvtt}
\end{table*}
\begin{table*}[t!]\centering
%
\subfloat[VATEX. 
\label{table:vatex}]{\tablestyle{4pt}{1.1}
\def\w{12pt} 
\def\wmod{18pt} 
\def\wmethod{53pt} 
 \hspace{-5mm}
\resizebox{!}{60pt}{
\begin{tabular}{L{\wmethod}C{\wmod}C{\w}C{\w}C{\w}C{\w}}
        \shline
        ~ & ~ & \multicolumn{4}{c}{VATEX}\\
        \cmidrule{3-6}  Method & Mod. & B4 & R & M & C \\
        \hline
        \textsc{VaTeX}~\cite{wang2019vatex} & V & 28.4 & 47.0 & 21.7 & 45.1 \\
         ORG-TRL~\cite{Zhang_2020_CVPR} & V & 32.1 & 48.9 & 22.2 & 49.7 \\
         Support-set~\cite{patrick2020support} & V & 32.8 & 49.1 & 24.4 & 51.2 \\
         \demph{Support-set~\cite{patrick2020support}} & \demph{V} & \demph{32.5} & \demph{48.9} & \demph{24.1} & \demph{50.5} \\
         OpenBook~\cite{zhang2021open} & V+T & 33.9 & 50.2 & 23.7 & 57.5 \\
        \demph{\textsc{Value}~\cite{VALUE}} & \demph{V+T} & \demph{-} & \demph{-} & \demph{-} & \demph{58.1} \\
        \hline
        \textsc{SwinBERT} & V & \textbf{38.7} & \textbf{53.2} & \textbf{26.2} & \textbf{73.0} \\
        \shline
        
    \multicolumn{3}{c}{} \\ 
    \multicolumn{3}{c}{} \\ 
    \multicolumn{3}{c}{} \\ 
    \end{tabular}
}
}
\hfill
\subfloat[TVC.    \label{table:tvc}]{\tablestyle{4pt}{1.1}
\def\w{12pt} 
\def\wmod{18pt} 
\def\wmethod{36pt} 
\resizebox{!}{60pt}{
\begin{tabular}{L{\wmethod}C{\wmod}C{\w}C{\w}C{\w}C{\w}}
        \shline
        ~ & ~ & \multicolumn{4}{c}{TVC}\\
        \cmidrule{3-6}  Method & Mod.  & B4 & R & M & C \\
        \hline
        MMT~\cite{lei2020tvr} & V & 9.9 & 30.4 & 15.2 & 36.0 \\
        MMT~\cite{lei2020tvr} & T & 6.3 & 7.7 & 13.9 & 33.7 \\
        MMT~\cite{lei2020tvr} & V+T & 10.8 & 32.8 & 16.9 & 45.3 \\
        \demph{HERO~\cite{li2020hero}} & \demph{V+T} & \demph{12.3} & \demph{34.1} & \demph{17.6} & \demph{49.9} \\
        \demph{\textsc{Value}~\cite{VALUE}} & \demph{V+T} & \demph{11.6} & \demph{33.9} & \demph{17.6} & \demph{50.5} \\
        \hline
        \textsc{SwinBERT} & V & \textbf{14.5} & \textbf{36.1} & \textbf{18.5} & \textbf{55.4} \\
       \shline
    \multicolumn{3}{c}{} \\ 
    \multicolumn{3}{c}{} \\ 
    \multicolumn{3}{c}{} \\ 
    \multicolumn{3}{c}{} \\ 
    \end{tabular}
}
}
\hfill
\subfloat[YouCook2. 
\label{table:yc2}]{\tablestyle{4pt}{1.1}
\def\w{12pt} 
\def\wmod{18pt} 
\def\wmethod{65pt} 
\resizebox{!}{60pt}{
\begin{tabular}{L{\wmethod}C{\wmod}C{\w}C{\w}C{\w}C{\w}C{\wmod}}
        \shline
        ~ & ~ & \multicolumn{5}{c}{YouCook2}\\
        \cmidrule{3-7} Method & Mod. & 
        B3 & 
        B4 & M & R & C\\
        \hline
        Masked Trans.~\cite{zhou2018end} & V &
        7.5 &
        3.8 & 10.6 & - & 37.9 \\
        DPC~\cite{shi2019dense} & V  & 
        - &
        2.2 & 17.6 & - & - \\  
        DPC~\cite{shi2019dense} & V+T &
        - &
        2.8 & 18.1 & - & -  \\
        \demph{VideoBERT~\cite{sun2019videobert}} & \demph{V} & 
        \demph{7.5} & 
        \demph{4.3} & \demph{11.9} & \demph{-} & \demph{55.0} \\ 
        \demph{ActBERT~\cite{zhu2020actbert}} & \demph{V} &
        \demph{8.6} &
        \demph{5.4} & \demph{13.3} & \demph{-} & \demph{65.0} \\ 
        AT~\cite{shi2019dense} & T & 
        - &
        8.5 & 16.9 & - & 106.0 \\
        AT+Video~\cite{shi2019dense} & V+T &
        - &
        9.0 & 17.7 & - & 112.0 \\ 
        \textsc{Value}~\cite{VALUE} & V &
        - &
        - & - & - & 53.6 \\ 
        \demph{\textsc{Value}~\cite{VALUE}} & \demph{V+T} &
        \demph{-} &
        \demph{\textbf{12.4}} & \demph{\textbf{18.8}} & \demph{\textbf{40.4}} & \demph{\textbf{130.3}} \\ 
        \hline
        \textsc{SwinBERT} & V & 
        \textbf{13.8} & 
        \textbf{9.0} & \textbf{15.6} & \textbf{37.3}  & \textbf{109.0} \\
        \shline
    \end{tabular}
}
}

\caption{Comparison with state-of-the-art methods on YouCook2, TVC, and VATEX. We gray out models that adopt vision-and-language pre-training on large-scale datasets for a fair comparison.
\label{tab:tvc_vatex_youcook2}
}
\vspace{-4mm}
\end{table*}

\subsection{Experimental Setup}

\paragraph{Datasets.} We conduct experiments on 5 video captioning datasets, detailed below.
\vspace{-2mm}
\begin{itemize}[leftmargin=*]
\item \textbf{MSVD}\cite{chen-dolan-2011-collecting} is a collection of $2K$ open-domain video clips downloaded from YouTube. Each video clip has roughly $40$ ground-truth captions written by human. \kevinarxiv{Similar to the prior work~\cite{venugopalan2014translating}, we use the standard split which contains $1.2K$ training videos, $100$ validation videos, and $670$ test videos. We compare our results with prior studies on the test split, and use validation split for ablation study.}
\vspace{-2mm}
\item \textbf{YouCookII}\cite{zhou2018towards} is a cooking domain dataset covering $89$ recipes. There are $15.4K$ video clips, and each has $1$ ground-truth caption. We use the standard training/validation split in the experiments.
\vspace{-2mm}
\item \textbf{MSRVTT}\cite{xu2016msr} consists of $10K$ open-domain video clips. Each video clip has $20$ ground-truth captions. We use the standard captioning split, which has $6.5K$ training videos and $2.9K$ testing videos. \kevinarxiv{We compare our results with prior studies on the test split, and use validation split for ablation study.}
\vspace{-2mm}
\item \textbf{TVC}\cite{lei2020tvr} is a TV domain dataset. There is a total of $262K$ caption descriptions paired with $108K$ video segments. The captions in TVC not only describe the video contents, but it may also describe the subtitles.
\vspace{-2mm}
\item \textbf{VATEX}\cite{wang2019vatex} is a relative large open-domain dataset, which contains $41.3K$ videos. Each video clip has $20$ ground-truth captions. We use the official training set for training, and evaluate the results using the public test set. \kevinarxiv{In supplementary material, we present additional results on private test split, where the scores are obtained from \textsc{Value} leaderboard evaluation server~\cite{value-server}.}
\end{itemize}

\vspace{-2mm}
\paragraph{Implementation Details.} 
We implement our model using Pytorch~\cite{paszke2019pytorch}, Huggingface transformer~\cite{wolf2020transformers}, and DeepSpeed library~\cite{rasley2020deepspeed}. The VidSwin is initialized with Kinetics-600 pre-trained weights~\cite{liu2021video}, and the multimodal transformer encoder is randomly initialized. In order to ensure that the video tokens have the same embedding size as that of the word tokens, we transform the video tokens using a learnable MLP. Following~\cite{VALUE}, we employ AdamW optimizer~\cite{loshchilov2017decoupled} and use a learning rate warm-up during the early 10\% training steps followed by linear decay. Additional details can be found in the supplementary material.

\begin{table*}[t!]\centering
\subfloat[Impact of \#Video Frames (\textit{T}). 
\label{tbl:frame-num}]{\tablestyle{6pt}{1.1}
\def\w{15pt}  
\begin{tabular}{lcc}
    \toprule
	\textit{T} &  MSRVTT  & VATEX \\
	\midrule
	2  & $36.6$ & $47.4$\\
	4  & $43.7$ & $58.2$\\
    8  & $47.6$ & $65.2$\\
    16 & $49.5$ & $68.4$\\
    32 & $52.3$ & $71.1$\\
    64 & $\textbf{55.3}$ & $\textbf{72.7}$\\
	\bottomrule
\end{tabular}
}
\hfill
\subfloat[Learning of Sparse Attention Mask.    \label{tbl:ablation-sparse}]{\tablestyle{5pt}{1.1}
\def\w{15pt}  
\begin{tabular}{ccccc}
    \toprule
	\textit{T} & Learnable Att. Mask & $\mathcal{L}_{\textsc{sparse}}$ & MSRVTT  & VATEX  \\
    \midrule
    32 & \xmark  & \xmark & $52.3$ & $71.1$\\  
    32 & \cmark & \xmark & $53.3$ & $70.7$\\
    32 & \cmark & \cmark & $\textbf{55.1}$ & $\textbf{71.6}$\\
	\bottomrule
    \multicolumn{5}{c}{} \\ 
    \multicolumn{5}{c}{} \\ 
    \multicolumn{5}{c}{} \\ 
\end{tabular}
}
\hfill
\subfloat[Heuristic vs. Learnable Attention Mask 
\label{tbl:heuristic}]{\tablestyle{5pt}{1.1}
\def\w{15pt}  
\begin{tabular}{lcc}
    \toprule
	Att. Mask  & MSRVTT & VATEX  \\
	\midrule
	Full Attention  & $52.3$ & $71.1$\\
	Spatial Window  & $51.9$ & $71.0$\\
	Temporal Window  & $51.0$ & $70.2$\\
    Ours (Learnable, Sparse)  & $\textbf{55.1}$ & $\textbf{71.6}$\\
	\bottomrule
    \multicolumn{3}{c}{} \\ 
    \multicolumn{3}{c}{} \\ 
\end{tabular}
}
\vspace{-5mm}
\caption{Results on  \textsc{SwinBERT} with 
\textbf{(a)} varying number of video frames (without learnable sparse attention mask), \textbf{(b)} ablation study on learnable attention mask and sparse attention loss, and \textbf{(c)} comparisons between heuristic and learnable attention masks. All experiments are conducted with 32 frames unless specified otherwise. All results are reported on CIDEr metric.
\label{tab:ablation_msrvtt_vatex}
}
\vspace{-4mm}
\end{table*}

\subsection{Main Results}


We compare \textsc{SwinBERT} with previous state-of-the-art methods on 5 public benchmark datasets. Following the literature~\cite{VALUE,zhang2021open,Zhang_2020_CVPR,wang2019vatex}, we provide detailed comparisons using a diverse set of performance metrics, including BLEU4~\cite{papineni2002bleu}, METEOR~\cite{banerjee2005meteor}, ROUGE-L~\cite{lin2004automatic} and CIDEr~\cite{vedantam2015cider}.

Table~\ref{table:msvd-msrvtt} shows detailed comparisons on MSVD and MSRVTT datasets. \kevinarxiv{\textsc{SwinBERT} outperforms previous state-of-the-art methods in terms of CIDEr metric by a large margin. Specifically, \textsc{SwinBERT} brings significant CIDEr improvements on MSVD (\textit{i.e.}, $+25.4$ higher than the prior arts).} 

In Table~\ref{table:vatex}, we report detailed comparisons on the VATEX dataset. \textsc{SwinBERT} achieves better performance than the prior works, especially on CIDEr metric.
It should be noted that previous state-of-the-art methods (\textit{i.e.}, \textsc{Value}~\cite{VALUE} and Support-set~\cite{patrick2020support}) perform vision-and-language (VL) pre-training on large-scale datasets for improving multimodal representations, whereas the results with \textsc{SwinBERT} are not based on VL pre-training. We believe that further integration of VL pre-training will provide additional improvements. This, from another point of view, demonstrates the superior performance of \textsc{SwinBERT}.

We further conduct analysis on the challenging TVC dataset, and the results are shown in Table~\ref{table:tvc}. Note that captions in TVC are designed to describe not only the visual events but also supplementary information presented in the subtitle sentences. \textsc{Value}~\cite{VALUE}, HERO~\cite{li2020hero} and MMT~\cite{lei2020tvr} are three prior works, that leverage multimodal video inputs, including 2D/3D visual frame features and subtitle sentences from the original TV show scripts.
With video frame inputs alone, \textsc{SwinBERT} is able to achieve better performance than all three of them. This superior performance  suggests that \textsc{SwinBERT} is effective in exploiting visual representations for video captioning. 

Table~\ref{table:yc2} shows the detailed comparisons on YouCook2. 
We list the prior works that take visual and/or textual modality signals as inputs. 
Compared with visual-only approaches, \textsc{SwinBERT} brings significant CIDEr improvements on YouCook2. To be specific, \textsc{SwinBERT} achieves $109.0$ CIDEr score, which is $+55.4$ higher than that of \textsc{Value}~\cite{VALUE}, and $+44.0$ higher than that of ActBERT~\cite{zhu2020actbert}. We believe that \textsc{SwinBERT} can be further enhanced with multimodal video inputs by leveraging additional modalities such as subtitle and audio, which is worth exploring in future study.

\subsection{Ablation Study}
We conduct comprehensive ablation study on multiple datasets to investigate the capability of the proposed model. Following~\cite{VALUE}, we use CIDEr metric~\cite{vedantam2015cider} as our primary evaluation metric for video captioning.

\Paragraph{Impact of video frames.}We first investigate the impact of the sampling rate of the video frames on the task of video captioning. Specifically, we uniformly sample $T=\{2,4,8,16,32,64\}$ frames from the given video clip to train and test our \textsc{SwinBERT}.  For clarity, we disable the sparse attention mask in this experiment. Table~\ref{tbl:frame-num} shows the model performance with varying number of video frames on MSRVTT and VATEX. As we increase the number of frames, we observe consistent improvements on the CIDEr metric. These results suggest that the performance of video captioning can be greatly lifted by using more densely sampled frames. 

\begin{table}[t]
\tablestyle{4pt}{1.1} 
\def\w{20pt} 
\def\wlong{45pt} 
\def\wshort{5pt} 
\centering
\begin{tabular}{llcccccc}
    \toprule
	\textit{T} & \scriptsize{Attn. Mask}  & \scriptsize{MSVD} & \scriptsize{YouCook2}& \scriptsize{MSRVTT}& \scriptsize{TVC} & \scriptsize{VATEX} \\
    \midrule
    32 & \scriptsize{Full}  & $127.9$ & $104.2$ & $52.3$ & $53.0$ & $71.1$\\
    32 & \scriptsize{Sparse (soft)} & $\textbf{147.6}$ & $\textbf{104.8}$ & $55.1$ & $\textbf{53.8}$ & $\textbf{71.6}$\\
    32 & \scriptsize{Sparse (binary)}  & $141.0$ & $101.4$ 
    & $\textbf{55.3}$ & $52.8$ & $\textbf{71.6}$\\
    \midrule
    48 & \scriptsize{Full}  & $144.2$ & $103.1$ & $53.9$ & $53.9$ & $71.7$\\
    48 & \scriptsize{Sparse (soft)}  & $147.8$ & $\textbf{105.0}$ & $54.6$ & $\textbf{55.2}$ & $71.9$\\
    48 & \scriptsize{Sparse (binary)}   & $\textbf{148.1}$ & $103.8$ & $\textbf{54.9}$ & $52.6$ & $\textbf{72.0}$\\
    \midrule
    64 & \scriptsize{Full} & $144.7$ & $106.1$ & $55.3$ &  $54.3$ & $72.7$\\
    64 & \scriptsize{Sparse (soft)} & $\textbf{149.4}$ & $\textbf{109.0}$ & $\textbf{55.9}$ & $\textbf{55.4}$ & $\textbf{73.0}$ \\ 
    64 & \scriptsize{Sparse (binary)}   & $146.3$ & $106.6$ & $55.0$ & $53.1$ & $71.9$\\
	\bottomrule
\end{tabular}
\vspace{-2mm}
\caption{Effectiveness of soft/binary sparse attention mask on longer video sequences. \textit{T} indicates number of video frames. All results are reported on CIDEr metric. 
}
\vspace{-4mm}
\label{tbl:longer-seq}
\end{table}

\Paragraph{Effectiveness of sparse attention mask.}One important question is whether adding sparse attention mask to the transformer is helpful. To understand the effect of the sparse attention mask, Table~\ref{tbl:ablation-sparse} shows the ablation study. First of all, we present a baseline that does not have any learnable attention mask, shown in the first row of Table~\ref{tbl:ablation-sparse}. In the second row, we show  another baseline which uses a learnable attention mask but no sparsity constraints are added. This is equivalent to a random attention mask. Finally, the bottom row shows our proposed method. We observe that the proposed sparsity constraint is helpful in improving video captioning in terms of CIDEr scores (\textit{i.e.}, $+2.8$ on MSRVTT and $+0.5$ on VATEX). 

\begin{table*}[t!]\centering
\subfloat[Transfer between different frame rates within a dataset
\label{tbl:transfer-frame}]{\tablestyle{5pt}{1.1}
\def\w{15pt}  
\begin{tabular}{lccccc}
    \toprule
	\#Video Frames (\textit{T}) & VATEX  & MSVD  & YouCook2  & MSRVTT  & TVC  \\
    \midrule
    32 & $71.6$ & $147.6$ & $104.8$ & $55.1$ & $53.8$ \\
    64 & $\textbf{73.0}$ & $149.4$ & $\textbf{109.0}$ & $\textbf{55.9}$ & $55.4$ \\
    \specialcelll{32 $\rightarrow$ 64\\(attn. mask only)} & $\textbf{73.0}$ &  $150.0$ &  $108.2$ &  $\textbf{55.9}$ &  $\textbf{56.9}$\\ 
    \specialcelll{32 $\rightarrow$ 64\\(entire model)}& $72.4$ &  $\textbf{152.3}$ &  $106.8$ &  $55.6$ &  $56.0$ \\ 
	\bottomrule
\end{tabular}
}
\hfill
\subfloat[Transfer across datasets \label{tbl:transfer-dataset}]{\tablestyle{5pt}{1.1}
\def\w{15pt}  
\begin{tabular}[t]{lcc}
    \toprule
	\textit{Dataset} & CIDEr\\
    \midrule
    VATEX & $71.6$ \\
    MSVD & $147.6$ \\
    \specialcelll{VATEX $\rightarrow$ MSVD\\(att. mask only)} & $148.1$\\
    \specialcelll{VATEX $\rightarrow$ MSVD\\(entire model)}& $\textbf{160.3}$\\
	\bottomrule
\end{tabular}
\begin{tabular}[t]{ccc}

	\multicolumn{3}{c}{} \\ 
    \multicolumn{3}{c}{} \\ 
    \multicolumn{3}{c}{} \\ 
    \multicolumn{3}{c}{} \\ 
    \multicolumn{3}{c}{} \\ 
\end{tabular}
\begin{tabular}[t]{lcc}
    \toprule
	\textit{Dataset} & CIDEr\\
    \midrule
    VATEX & $71.6$ \\
    MSRVTT & $55.1$ \\
    \specialcelll{VATEX $\rightarrow$ MSRVTT\\(att. mask only)} & $\textbf{55.8}$\\
    \specialcelll{VATEX $\rightarrow$ MSRVTT\\(entire model)} & $54.5$\\
	\bottomrule
\end{tabular}
}
\vspace{-3mm}
\caption{Transferability of \textsc{SwinBERT} 
\textbf{(a)} between different frame rates within a dataset and \textbf{(b)} across datasets. We experiment with two settings: ($i$) transfer the learned sparse attention masks only (attn. mask only) and ($ii$) transfer the learned model weights along with sparse attention masks (entire model). All results are reported on CIDEr metric.
}
\vspace{-4mm}
\end{table*}

\Paragraph{Comparison between heuristic and learnable attention masks.}We also study the design of attention patterns for constructing our sparse attention mask. To be specific, we explore two heuristic designs including  \textit{(i) Spatial Window}: A sliding window attention pattern that attends to its neighbor tokens along the spatial dimension; \textit{(ii) Temporal Window}: A sliding window attention pattern, which attends along the temporal dimension. We use a fixed window size $w$ for both Spatial Window and Temporal Window, and we have explored $w=\{10,20,50,100\}$ in our experiments. 
Table~\ref{tbl:heuristic} presents results with different sparse attention masks along with a \textit{Full Attention} baseline, that is the original attention mask allowing full attentions among all video tokens. Results show that both Spatial Window and Temporal Window brings performance degradation, compared to the Full Attention baseline. In contrast, our learnable sparse attention mask improves over Full Attention and heuristic sparse attention masks.  We conjecture that our sparsity constraint enforces the model to identify more salient video frame patches along both spatial and temporal dimensions for caption generation. Visualizations of the learned sparse attention masks shown later in this section further corroborates our hypothesis. 


\Paragraph{Longer video sequences.}We further examine the capability of the proposed sparse attention mask using longer video sequences. We apply the learnable sparse attention masks to $T=\{32,48,64\}$ frames uniformly sampled from the video clips, and the results are shown in Table~\ref{tbl:longer-seq} (\textit{Full vs. Sparse (soft)}). We observe that adding sparse attention mask to \textsc{SwinBERT} consistently improves the CIDEr scores  across different video sequence length, and push the limit to new state-of-the-arts on all the 5 benchmarks. These results suggest that the sparse attention mask is effective in regularizing model training for long-range video sequence modeling.

\Paragraph{Binary sparse attention mask.}Our learnable sparse attention mask can be seen as a \textit{soft} attention mask, which consists of continuous values between $0$ and $1$. An interesting question we aim to answer is: \textit{can we enforce it into a binary mask?} We test this hypothesis by simply threshholding the learned sparse attention mask with a fixed threshold $0.5$. In addition, we fine-tune the model for a few training steps to adapt it to the binarized mask. In Table~\ref{tbl:longer-seq}, we observe that converting the mask to a binary one may have a slight performance drop on the CIDEr metric, which is expected as we reduce the capacity of the attention mask. It is worth noting that, with the binarized mask, the caption performance is comparable or better than the Full Attention (Full) baseline. In future, we plan to leverage custom CUDA implementations to construct this binary sparse attention mask to improve runtime speed.

\Paragraph{Generalization capability.}Since our sparse attention mask is optimized for task-specific performance improvements, one may wonder its generalizabilty to different frame rates and different datasets. We study the generalization capability under two configurations: \textit{(i) Across frame rates}: we first train \textsc{SwinBERT} at a slow frame rate, and then move to a faster frame rate for further training. To achieve this, we expand the learned sparse attention mask by linear interpolation along the temporal dimension; \textit{(ii) Across datasets}: we first train \textsc{SwinBERT} on one dataset, and then fine-tune it on another dataset. The experiments are conducted in two settings, transferring the whole model weights or only the sparse attention mask. 

Table~\ref{tbl:transfer-frame} shows the results of transferring from $32$ frames to $64$ frames. We observe that it yields a comparable or better CIDEr score compared to using $64$ frames directly. It should be noted that, transferring only the sparse attention mask is able to achieve reasonable CIDEr scores on the 5 datasets. The results suggest that linear interpolation along the temporal dimension is effective for transferring between different frame rates. 



Table~\ref{tbl:transfer-dataset} shows the results of transferring across datasets. In this experiment, we first train our model on VATEX dataset, and then fine-tune it on MSRVTT and MSVD datasets, respectively. We observe that such transfer learning scheme improves CIDEr scores for both datasets. 
As the data domain between VATEX and MSVD is similar, fine-tuning the entire model is more effective for improving CIDEr scores on MSVD.
The success in transfer learning suggests that the performance of \textsc{SwinBERT} can be further improved with pre-training on even larger-scale video-text datasets, which we leave as future study.

\begin{figure*}[t!]
\begin{center}
\includegraphics[trim=0 0 0 0, clip,width=0.98\textwidth]{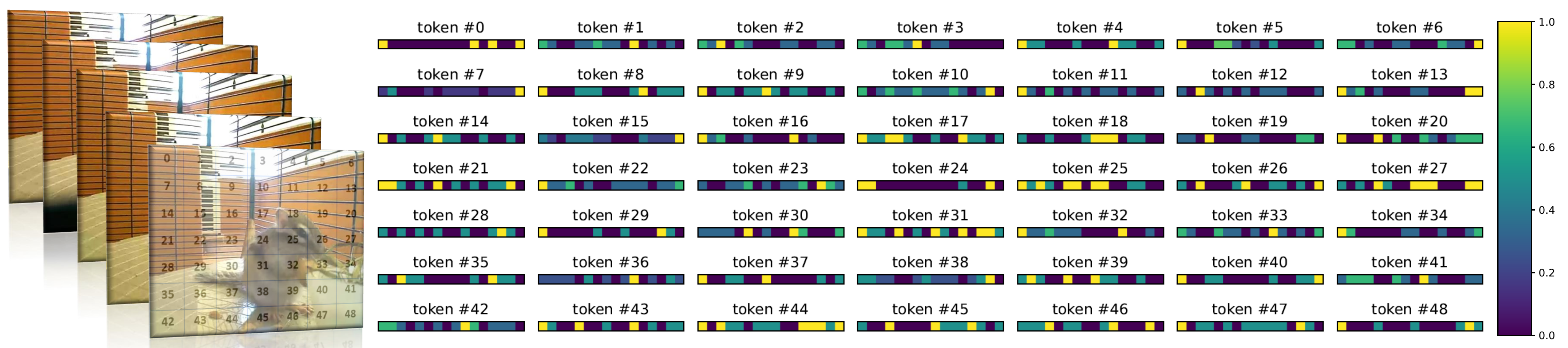}
\vspace{-1mm}
\caption{
\textbf{Visualization of sparse attention mask along the temporal dimension.} Our sparse attention mask discovers possible principle in the video sequences. We observe that boundary-region tokens can be sparsely sampled along the temporal dimension. This is probably due to similar background in a video clip. On the other hand, as the center-region tokens may contain more pixel variations (such as movements, actions, or scene changes), they thus require denser sampling along the temporal dimension.}
\label{fig:temporal}
\end{center}
\vspace{-6mm}
\end{figure*}

\Paragraph{Visualization of sparse attention mask.}
We visualize the learned sparse attention pattern in Figure~\ref{fig:temporal}. Note that the values are obtained from the \textit{soft} attention mask without thresholding.
On the left, we show an example video clip which is randomly sampled from MSVD dataset. Additionally, we denote the patch regions and the corresponding token IDs at the first frame. On the right, for each token, we visualize the weights of the learned sparse attention mask along the temporal dimension using a horizontal bar, where yellow color indicates stronger attention activity. We briefly summarize our findings: \textit{(i)} Many of the tokens at the boundary are attending to some starting and ending frames. This is possibly because the background does not change much, and therefore for those tokens, the temporal information can be sparsely sampled with respect to the attention mask; \textit{(ii)} The center-region tokens may contain more movements or scene changes, therefore require denser sampling along the temporal dimension.



\begin{figure}
\centering
 \subfloat[(a)][Sparsity of the attention mask\label{fig:sparsity}]{
   \includegraphics[trim=0 0 0 0, clip,width=0.48\linewidth]{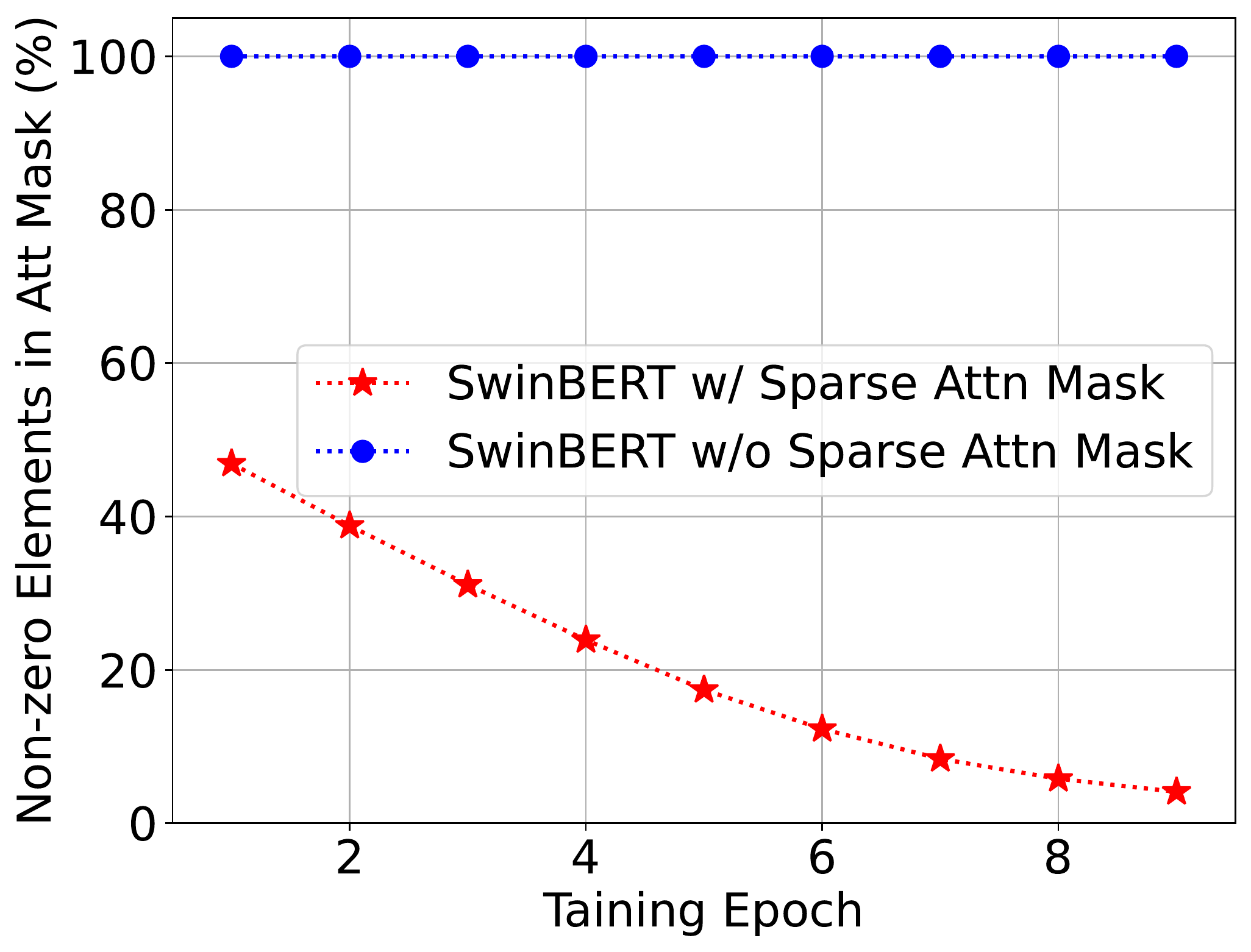}
 }
 \subfloat[(b)][CIDEr score\label{fig:cider}]{
   \includegraphics[trim=0 0 0 0, clip,width=0.48\linewidth]{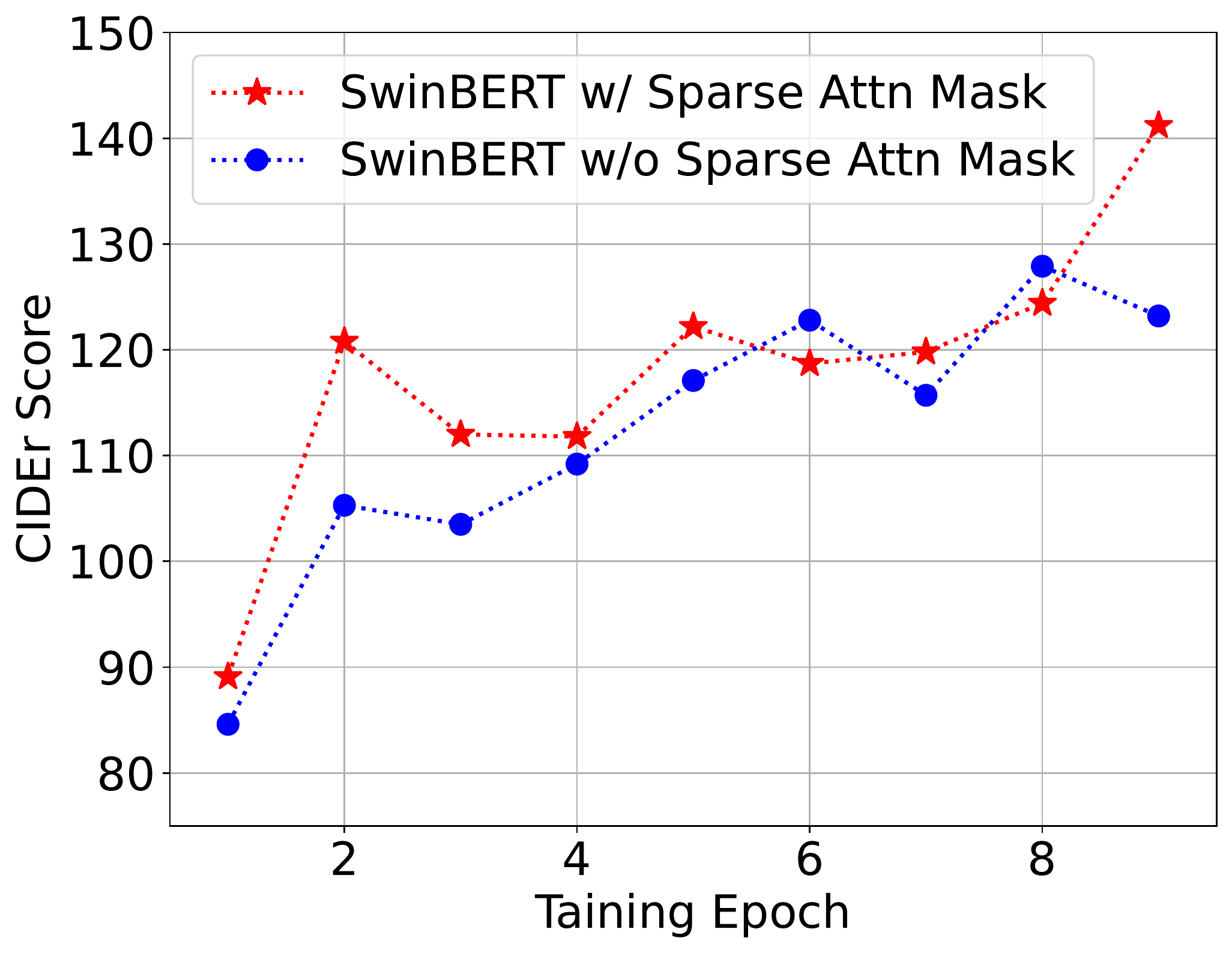}
 }
\vspace{-2mm}
\caption{\small
Training behavior of \textsc{SwinBERT}. \textbf{(a)} During training, the proposed sparsity constraint effectively reduces the percentage of non-zero elements in the attention mask. \textbf{(b)} Sparsity constraint does not interfere captioning as CIDEr score keeps increasing. }
\vspace{-6mm}
\label{fig:sparse}
\end{figure}

\Paragraph{Training behavior.}In Figure~\ref{fig:sparse}, we investigate the learning behavior of our sparse attention mask. In Figure~\ref{fig:sparsity}, our proposed sparsity constraint is effective in reducing the number of non-zero elements in the attention mask, and more than $95\%$ of the elements are set to zero in the end. This verifies the sparsity of the attention mask. In addition, as shown in Figure~\ref{fig:cider}, we find that our sparsity constraint does not interfere the learning of video captioning, as CIDEr scores keep increasing during the learning process.

\begin{figure}
\begin{center}
\vspace{-1mm}
\includegraphics[trim=0 0 0 0, clip,width=1.0\linewidth]{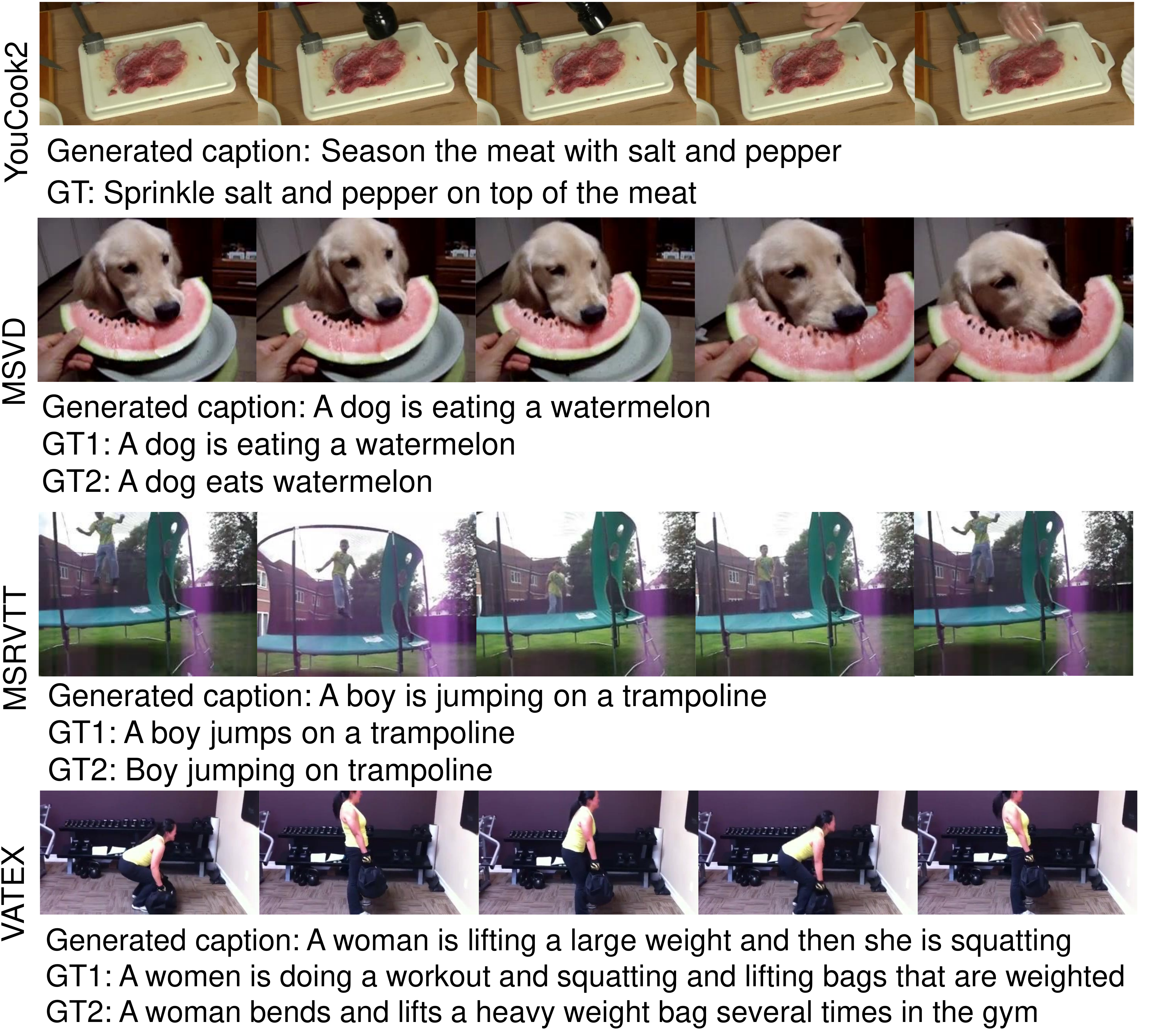}
\vspace{-6mm}
\caption{\small
Qualitative examples generated by \textsc{SwinBERT}. The generated captions are semantically reasonable and describe the video contents correctly.}
\vspace{-10mm}
\label{fig:caption}
\end{center}
\end{figure}

\Paragraph{Qualitative results.}Figure~\ref{fig:caption} shows the qualitative examples of \textsc{SwinBERT}. We find that 
\textsc{SwinBERT} is capable of recognizing the visual contents (\textit{e.g.}, dog and watermelon), and correctly describes the actions and events (\textit{e.g.}, eating) in the given video. We also note that, while our model generates semantically reasonable captions, the predicted word sequences may not always equal to the ground truth.   

\section{Conclusion}\label{sec:conclude}
We present \textsc{SwinBERT}, a new end-to-end fully Transformer-based architecture for video captioning. We further propose to adaptively learn a sparse attention mask for better video sequence modeling. Extensive experimental results on 5 popular benchmark datasets show that \textsc{SwinBERT} achieves better performance than the previous state-of-the-art methods by a large margin. In future, we plan to investigate large-scale video-language pre-training to further enhance the captioning performance. 






\section{Acknowledgements}
We thank Jianfeng Wang, Xiaowei Hu, Lin Liang, Zhengyuan Yang, Ehsan Azarnasab, Yue Cao, Lei Ji, Huaishao Luo and Ze Liu for the valuable discussions.

{\small
\bibliographystyle{ieee_fullname}
\bibliography{vidcap}
}

\clearpage
\appendix
\pdfoutput=1
\twocolumn[{%
\renewcommand\twocolumn[1][]{#1}%
\begin{center}
\textbf{\Large Supplementary Material}
\end{center}
\vspace{2mm}
}]

\begin{table*}
\centering
\begin{tabular}{lc cccc}
    \toprule
	Backbone & \#frames (\#tokens) & Attn. Mask & MSRVTT & MSVD & VATEX \\
	\midrule
	SOTA & - & - & 52.9~\cite{zhang2021open} & 95.2~\cite{Zhang_2020_CVPR} & 58.1~\cite{VALUE} \\
	\midrule
	TimeSformer & 8 (1568)& Full & 49.9 & 123.4 & 57.9 \\
	TimeSformer & 8 (1568)& Sparse & 51.9 & 127.6 & 63.0 \\
	VidSwin & 32 (784)& Full & 52.3 & 127.9 & 71.1 \\
	VidSwin & 32 (784)& Sparse & 55.1 & 147.6 & 71.6 \\
	\bottomrule
\end{tabular}
\caption{Analysis of our method with different video backbones. All backbones are pretrained on Kinetics-600~\cite{kinetics600}. We report CIDEr score~\cite{vedantam2015cider} in this analysis.}
\label{tbl:backbones}
\end{table*}

\begin{table*}
\centering
\begin{tabular}{lccc}
    \toprule
	Method & Backbone & Pretraining data for backbone & CIDEr \\
	\midrule
	VALUE~\cite{VALUE} & SlowFast & K400 & 51.2 \\Ours & SlowFast &  K400 & 53.6 \\
	
	\midrule
    VALUE~\cite{VALUE} & CLIP-ViT + SlowFast & 400M image-text pairs + K400 & 58.1 \\
	Ours & VidSwin &  ImageNet + K400 & 68.1 \\
	Ours & VidSwin &  ImageNet + K600 & 71.1 \\
	\bottomrule
\end{tabular}
\caption{Breakdown of pre-training data, evaluated on VATEX.}
\label{tbl:pretrain}
\end{table*}

\begin{table}
\centering
\begin{tabular}{lccc}
    \toprule
	Dataset & Val split & Test split & Private test split \\
	\midrule
	VATEX & 84.4 & 73.0 & 74.3 \\
	MSRVTT & 55.1 & 53.8 & - \\
	MSVD & 160 & 120.6 & - \\
	TVC & 57.0 & - & 49.7 \\
	YouCook2 & 109 & - & 101.3 \\
	\bottomrule
\end{tabular}
\caption{\kevinarxiv{Additional results on different splits. All results are reported on CIDEr metric.}}
\label{tbl:moreresults}
\end{table}

\section{Analysis of Different Video Backbones}

Table~\ref{tbl:backbones} shows our proposed method is generalizable to different video backbones. Note that the SOTA methods are trained with pre-extracted 2D and 3D CNN features. Our end-to-end trained model with only video backbone (TimeSformer~\cite{bertasius2021space} or Video Swin Transformer~\cite{liu2021video}), can often outperform the recent SOTA. Adding sparse attention mask consistently improves model performance across the video backbones considered. Further, a stronger backbone yields better captioning performance. 

It is worth noting that, TimeSformer generates longer video tokens compared to that of Video Swin Transformer (VidSwin). This introduces extra memory cost for the language model (due to quadratic complexity), making TimeSformer difficult to scale to longer sequences. Due to GPU memory constraints, during rebuttal period, we can only train TimeSfomer on 8 frames per clip.
From another perspective, this shows VidSwin offers a favorable memory-accuracy trade-off for video captioning.

\section{Influence of Pre-Training on Backbone}  
The top rows of Table~\ref{tbl:pretrain} give a fair comparison where both approaches use the same SlowFast~\cite{feichtenhofer2019slowfast} as the backbone. Our method achieves better performance than VALUE~\cite{VALUE}. 

The bottom rows of Table~\ref{tbl:pretrain} show the best results obtained by the two methods with different pre-training datasets. VALUE uses both CLIP-ViT~\cite{pmlr-v139-radford21a} and SlowFast~\cite{feichtenhofer2019slowfast} as backbones, which are pre-trained on 400M image-text pairs~\cite{pmlr-v139-radford21a} and Kinetics-400 (K400)~\cite{kinetics400}. In contrast, our video backbone is pre-trained on ImageNet~\cite{ILSVRC15} and K400/600. Although our video backbone uses less pre-training data than VALUE, we achieve better caption performance. We show that end-to-end training (from video patches to textual outputs) is crucial to the performance of video captioning. Compared with K400, pre-training backbone with K600 slightly improves CIDEr.

\begin{table*}
\centering
\begin{tabular}{lccccccc}
    \toprule
	 & 0 & 0.1 & 0.5 & 1 & 2 & 5 & 10 \\
	 \midrule
	 MSRVTT (32frm)& 52.3 & 53.4 & 53.8 & 53.9 & 54.9 & \textbf{55.1} & 53.4\\
	\bottomrule
\end{tabular}
\caption{Our model (with sparse attention) gives consistent improvements over baseline (without sparse attention) across different choices of $\lambda$.}
\label{tbl:lambda}
\end{table*}

\section{Choice of Hyperparameter $\lambda$} 

Since we use a regularization hyperparameter $\lambda$ in our sparsity constraint (see Eq.~\textcolor{red}{1} in our main manuscript), we provide further experiments with difference choices of $\lambda$. Table~\ref{tbl:lambda} shows that our model gives consistent improvements over different choices of $\lambda$.

\section{Additional Qualitative Results}

\begin{figure*}
\begin{center}
\vspace{-1mm}
\includegraphics[trim=0 0 0 0, clip,width=1.0\linewidth]{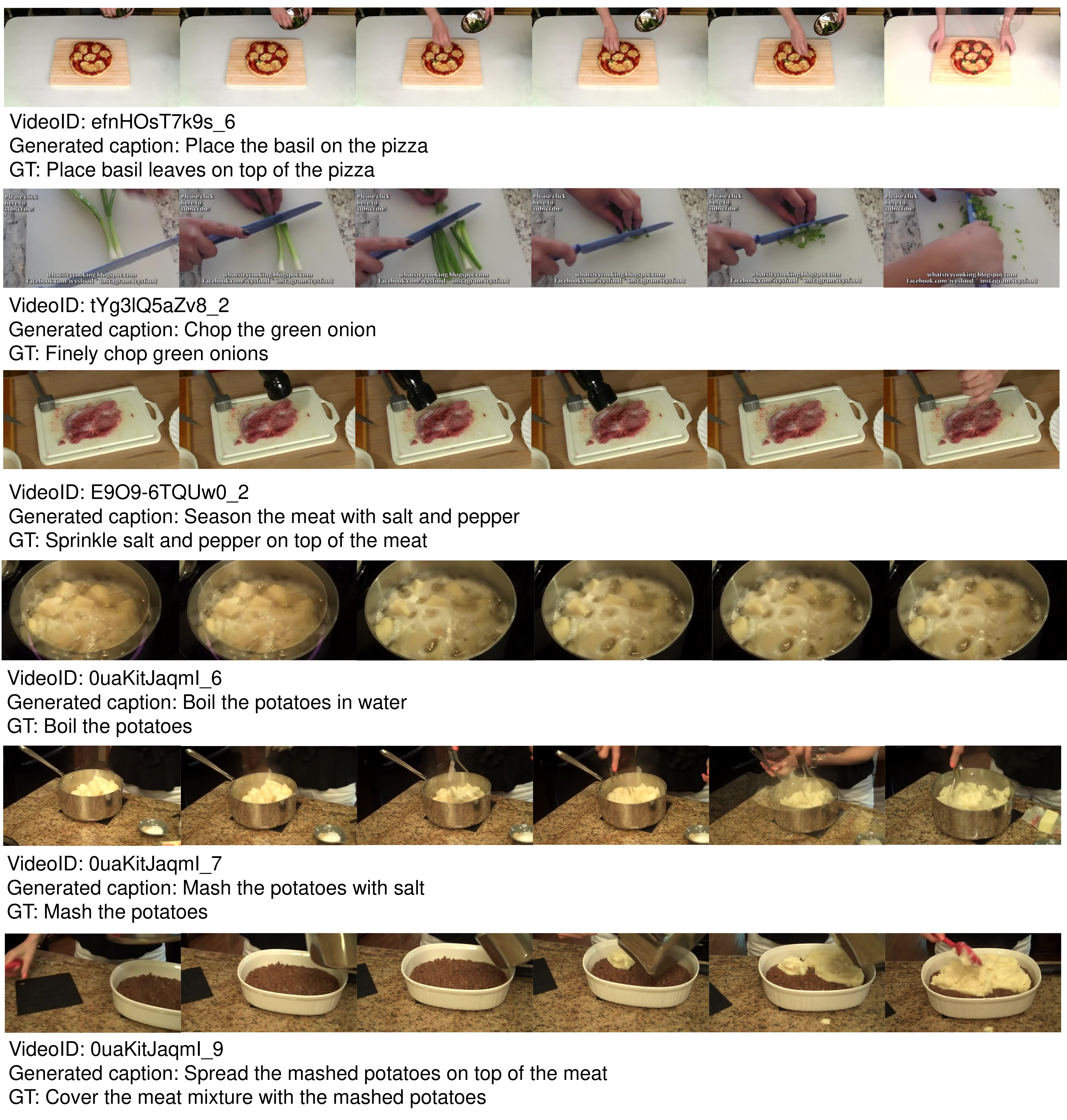}
\vspace{-7mm}
\caption{\small
Qualitative examples generated by \textsc{SwinBERT} on YouCook2 dataset.}
\vspace{-8mm}
\label{fig:qual-yc2}
\end{center}
\end{figure*}

\begin{figure*}
\begin{center}
\vspace{-1mm}
\includegraphics[trim=0 0 0 0, clip,width=1.0\linewidth]{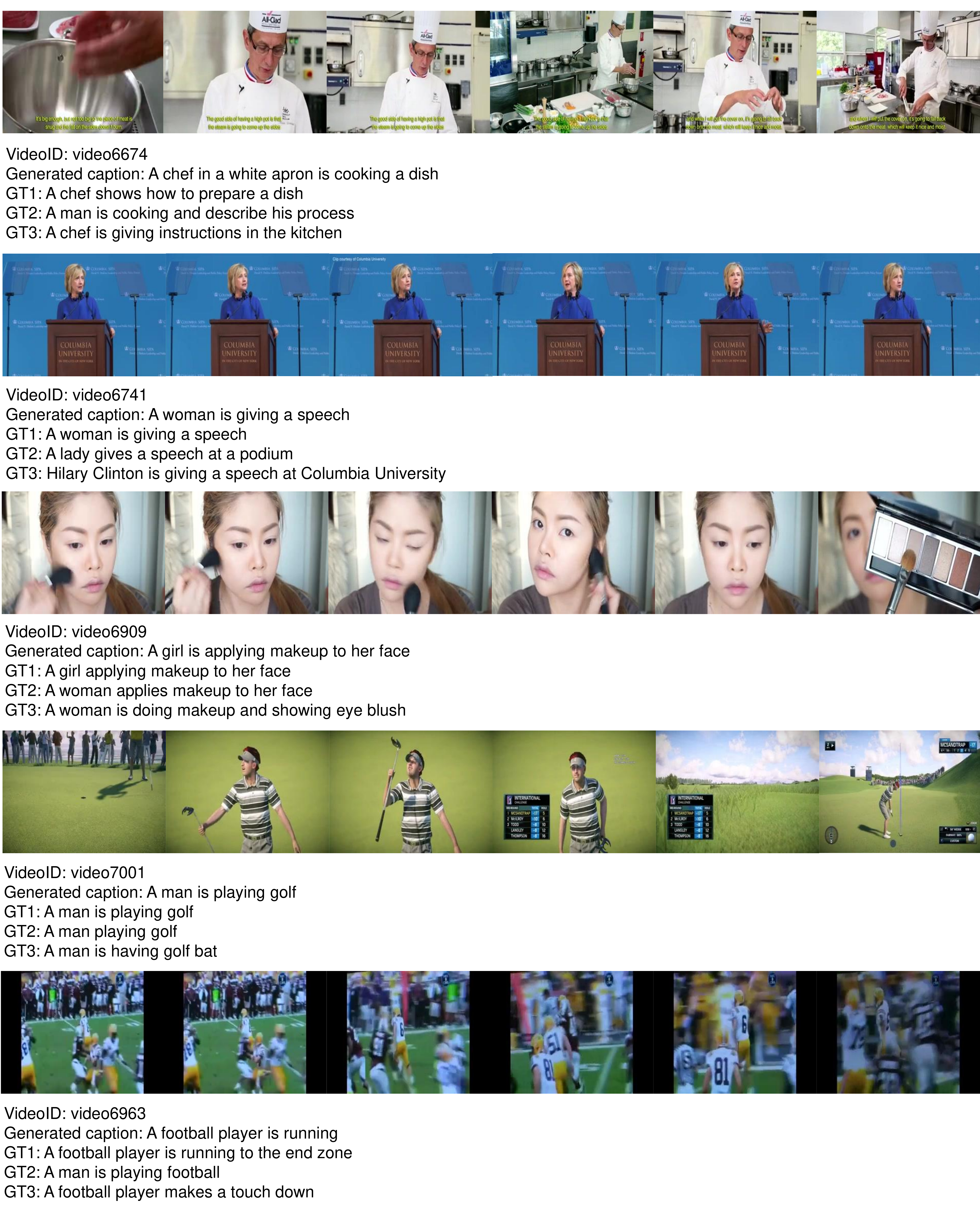}
\vspace{-7mm}
\caption{\small
Qualitative examples generated by \textsc{SwinBERT} on MSRVTT dataset.}
\vspace{-8mm}
\label{fig:qual-msrvtt}
\end{center}
\end{figure*}

\begin{figure*}
\begin{center}
\vspace{-1mm}
\includegraphics[trim=0 0 0 0, clip,width=1.0\linewidth]{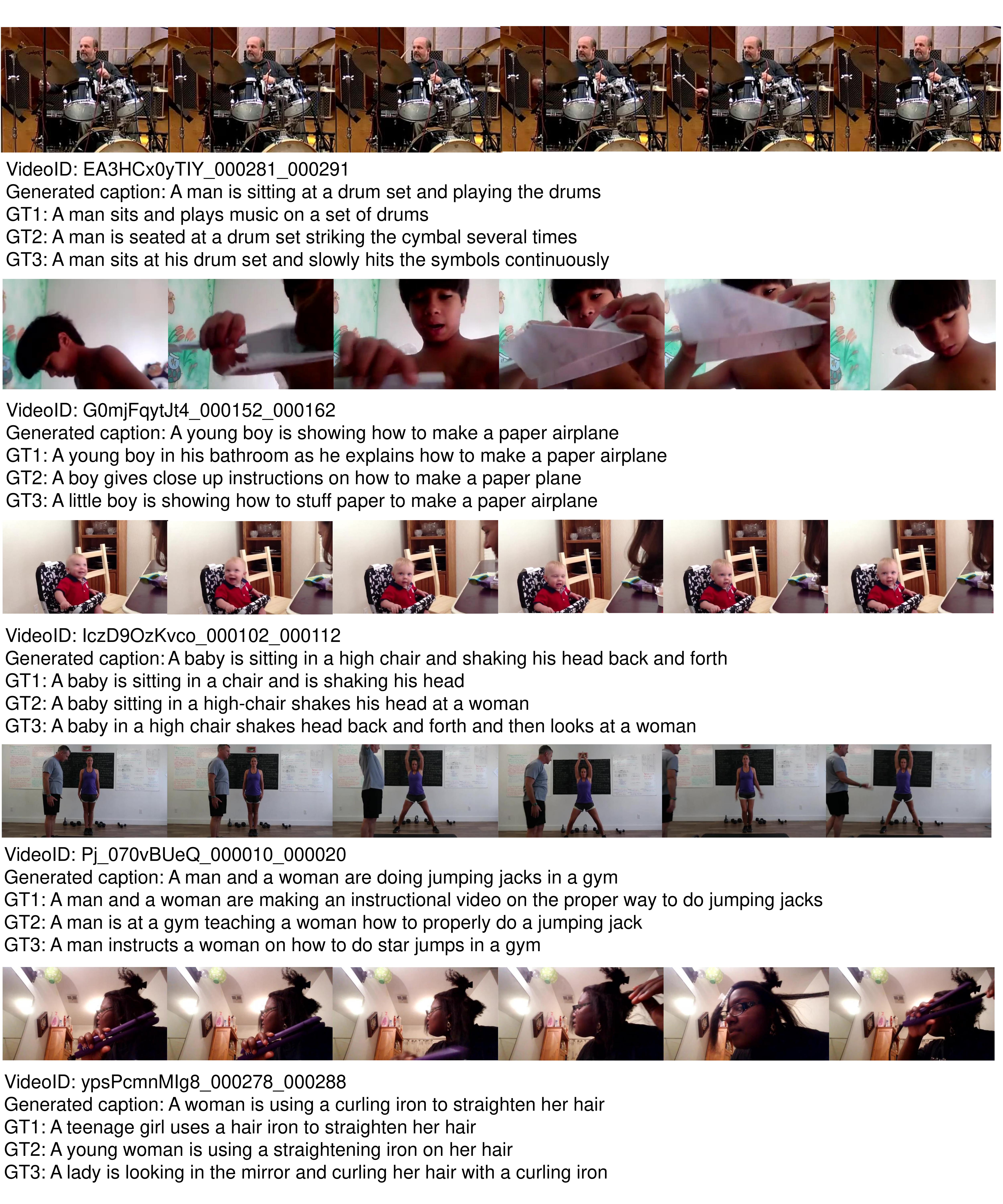}
\vspace{-7mm}
\caption{\small
Qualitative examples generated by \textsc{SwinBERT} on VATEX dataset.}
\vspace{-8mm}
\label{fig:qual-vatex}
\end{center}
\end{figure*}

\begin{figure*}
\begin{center}
\vspace{-1mm}
\includegraphics[trim=0 0 0 0, clip,width=1.0\linewidth]{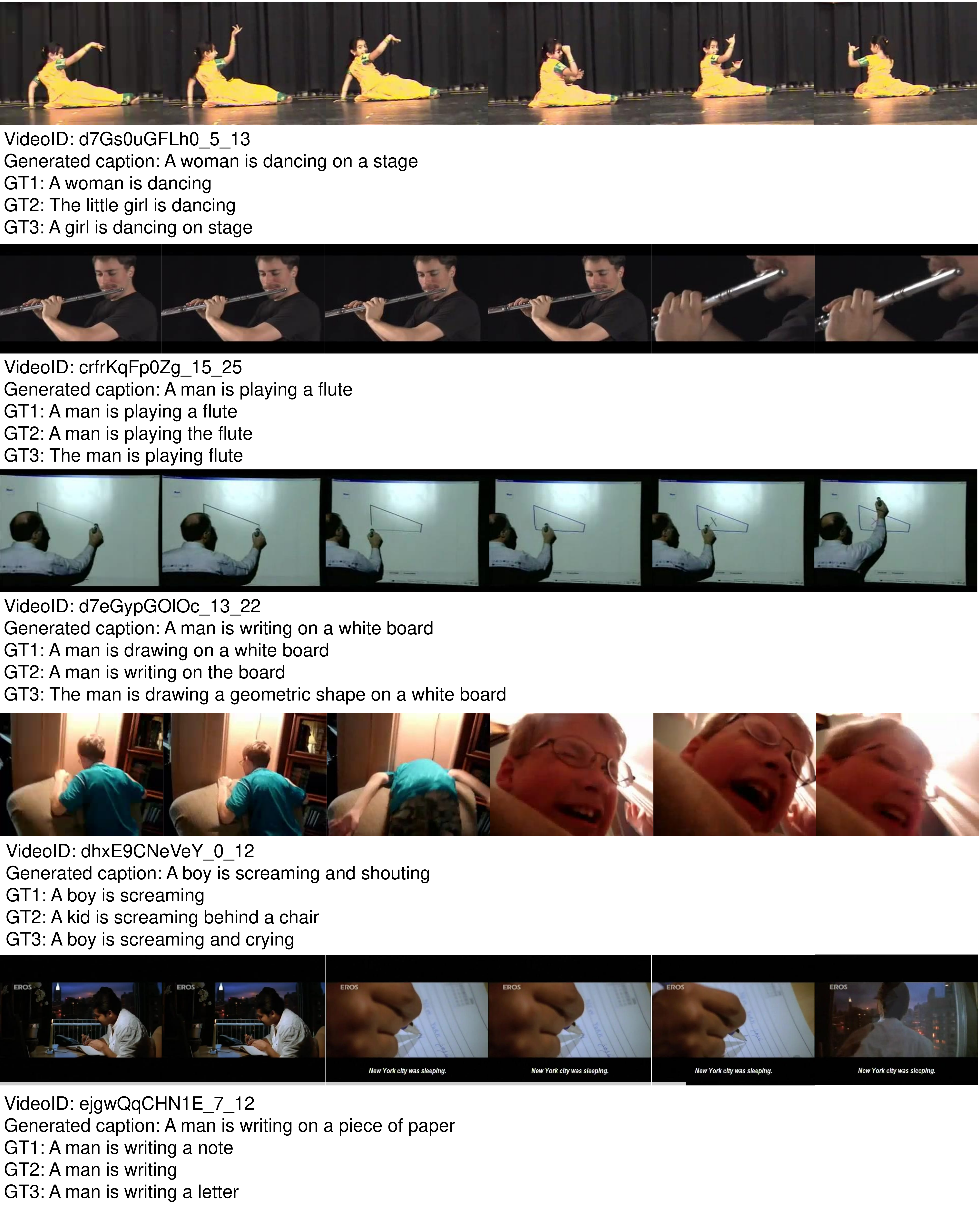}
\vspace{-7mm}
\caption{\small
Qualitative examples generated by \textsc{SwinBERT} on MSVD dataset.}
\vspace{-8mm}
\label{fig:qual-msvd}
\end{center}
\end{figure*}

We present additional qualitative results in Figure~\ref{fig:qual-yc2}, \ref{fig:qual-msrvtt}, \ref{fig:qual-vatex}, and \ref{fig:qual-msvd}. For each video, we show our prediction and the corresponding ground-truth captions. 

In Figure~\ref{fig:qual-yc2}, \textsc{SwinBERT} generates semantically correct captions for the considered cooking videos. For example, as presented in the top row, our model predicts ``\textit{Place the basil on the pizza,}" while the ground truth is ``\textit{Place basil leaves on top of the pizza}." Although the word sequences are not exactly the same, both can be considered semantically correct with respect to the given video.

Figure~\ref{fig:qual-msrvtt} shows our qualitative results on MSRVTT. We observe that \textsc{SwinBERT} works well for open-domain videos. For example, our model is capable of recognizing different actions, such as \textit{giving a speech}, \textit{applying makeup}, and \textit{playing golf}. In addition, some of our predictions are similar to the ground truths, as presented in the second, third, and fourth rows. 

In Figure~\ref{fig:qual-vatex}, we show our results on VATEX, where the ground-truth sentences are more descriptive and challenging. \textsc{SwinBERT} recognizes fine-grained objects (\textit{e.g.}, drum set, paper airplane, high chair, and curling iron) in various viewpoints, and generates semantically reasonable captions for the input videos.

Figure~\ref{fig:qual-msvd} shows the results on MSVD. \textsc{SwinBERT} recognizes the video events correctly. As presented in the first row, \textsc{SwinBERT} recognizes ``\textit{A woman is dancing on a stage}" by seeing detailed movements of the posture in multiple frames. In the second row, \textsc{SwinBERT} correctly describes ``\textit{A man is playing a flute}."


\section{Additional Training Details}

We implement our models based on PyTorch~\cite{paszke2019pytorch}. We also adopt mixed-precision training. To be specific, we use DeepSpeed~\cite{rasley2020deepspeed} for the majority of our experiments. Additionally, we use Nvidia Apex~\cite{nvidia-apex} for the experiments of longer video sequences, which empirically leads to more stable training. All experiments are conducted on Microsoft Azure~\cite{msft-azure} with multiple Nvidia V100 GPUs (32GB).

Our Video Swin Transformer (VidSwin) is a {\fontfamily{cmtt}\selectfont Swin-base} model initialized with Kinetics-600 pre-trained weights~\cite{liu2021video}. Our multimodal transformer has $12$ layers, and the hidden size is $512$. Our multimodal transformer is randomly initialized. Both VidSwin and the multimodal transformer are trained in an end-to-end manner. 

We resize the shorter side of all the video frames to $224$. During training, we random crop $(224\times224)$ at the same location for all the frames in a given video. During inference, we center crop $(224\times224)$ for all the frames.

Since the considered datasets have different data scales and domains, we use task-specific training epochs and learning rates based on the performance of validation sets. 


\section{Broader Impact and Ethical Concerns}

Video captioning offers the possibility to make videos more accessible and inclusive to all users, including low-vision and blind users~\cite{morris2020ai}. 
In this paper, we aim to improve the accuracy of video captioning with better video representations.
While our method outperforms the previous state-of-the-arts, the model does not always guarantee a perfect prediction. As a data-driven system, our model is sensitive to the distribution of training data, therefore may fail when encountering videos in the wild.  To avoid any undesirable predictions that could lead to ethical concerns in real-world applications (\textit{e.g.}, incorrect semantics, wrong identity), the generated caption should be considered as a draft that requires further editing. 





\section{\kevinarxiv{Additional Results on Different Splits}}
\kevinarxiv{In Table~\ref{tbl:moreresults}, we report performance of our model on both validation and test splits. In addition, we report our results on private test splits, where the scores are obtained from \textsc{Value} leaderboard evaluation server~\cite{value-server}.}

\section{Discussion}

\Paragraph{Computational Cost:} In this work, we primarily focus on improving caption accuracy (CIDEr score), and the sparse attention mask is used as a regularizer for improving training. Since we implement the sparse attention mask via an additional learnable embedding, it does not have a real speed-up. In the future, we plan to investigate CUDA implementations to construct a binary attention mask to reduce computational cost. In our current implementation, our model is computational memory intensive since both VidSwin and BERT require sufficient GPU memory during training. We use mix-precision and checkpointing to remedy the memory issues.

\Paragraph{How many frames are sufficient for video captioning:} Our experimental results in Table 4(a) of the main text suggest that more frames would benefit captioning performance. However, due to GPU memory constraints, with 128-frame inputs, we are restricted to use batch size=1, making the training inefficient. Hence, we can only empirically conclude that 64 frames give the best performance. Please note that 128-frame is a significant departure from current SOTA, which are typically 8-32 frames.

\Paragraph{Token selection:} Recently, researchers~\cite{xu2021evo} are exploring dynamic token selection to reduce the computation complexity of the transformer. While dynamic token selection is useful for vision or NLP transformers, it needs to be studied further when integrated with multimodal transformers for video captioning. 
Unlike previous efforts that attempted to reduce the number of tokens, we keep video tokens intact and improve caption accuracy by regularizing attention over time.

\Paragraph{Observation in VATEX and MSVD:} We observe the two datasets have different characteristics. The groundtruth captions in VATEX include detailed actions, and the caption model requires more temporal features to have a correct generation. For MSVD, the groundtruth captions are more about the scenes and objects, and thus spatial features play a critical role to captioning.

\end{document}